\documentclass[11pt]{article}

\usepackage[preprint]{acl}

\usepackage[T1]{fontenc}
\usepackage{times}
\usepackage{latexsym}
\usepackage[utf8]{inputenc}
\usepackage{graphicx}
\usepackage{algorithm}
\usepackage{algpseudocode}
\usepackage{amsmath} 
\usepackage{multicol}
\usepackage{subfigure}
\usepackage{booktabs}  
\usepackage{multirow}  
\usepackage{array}  
\usepackage{xspace}
\usepackage[hide]{todo}
\usepackage[normalem]{ulem}
\usepackage{inconsolata}
\usepackage{microtype}
\usepackage{enumitem}

\newcommand{\vocab}{\ensuremath{\mathcal{U}}}

\newcommand{\corpus}{\ensuremath{\mathcal{D}}}
\newcommand{\vtheta}{\ensuremath{\boldsymbol{\theta}}}
\newcommand{\defun}{\ensuremath{\triangleq}}
\title{Rethinking BPE: Merge Selection as Model Selection}

\author{Vijini Liyanage \and François Yvon \\
  Sorbonne-Université, CNRS, ISIR, Paris \\
\texttt{\{pilanaliyanage,yvon\}@isir.upmc.fr}
  }

\begin{document}
\maketitle
\begin{abstract}
Subword tokenization methods such as Byte-Pair Encoding (BPE) are a core component of modern language models; yet, tokenizer capacity remains typically determined by a fixed merge budget, chosen heuristically. Since larger vocabularies increase memory usage and computational cost, selecting compact and effective token inventories is essential for efficient language modeling. We study the optimization of a BPE merge budget as a statistical model selection problem and investigate information-theoretic criteria for \emph{computing optimal merge sets} directly from corpus statistics. To further improve the merge selection in limited data conditions, we augment the BPE training corpora with synthetic texts generated from the original data, providing estimates for the unigram and bigram distributions. Experiments across multiple corpora and downstream tasks show that using a principled vocabulary selection technique reduces model size, with little degradation in performance. We further observe that synthetic data can yield better tokenizers, especially in low-resource settings.
\end{abstract}


\section{Introduction} \label{sec:introduction}

Modern language models rely on subword tokenization to represent text efficiently. Methods such as Byte-Pair Encoding (BPE) \citep{gage-1994-anewalgorithm, sennrich-etal-2016-neural} construct vocabularies by iteratively merging frequent symbol pairs, balancing vocabulary size and sequence length. Since tokenization defines the units processed by a model, it affects compression efficiency, computational cost, and downstream performance. However, tokenizer design remains largely heuristic: most systems fix the number of BPE merges in advance, typically ranging from tens to hundreds of thousands depending on the model and linguistic coverage \citep{brown-2020-language, touvron-etal-2023-llama, bai-2023-qwen}. These choices are rarely principled and are often inherited from prior work or selected through parameter search.

The number of merges controls tokenization granularity and the tradeoff between vocabulary size and sequence length \citep{kudo-2018-subword}. Too few merges create fragmented representations and long sequences, while too many produce large vocabularies capturing spurious or domain-specific patterns. This tradeoff affects efficiency, as longer sequences increase transformer attention costs, whereas large vocabularies increase embedding and decoding overhead.

In this work, we investigate principled BPE merge selection from corpus statistics. We formulate merge selection as a model selection problem and explore information-theoretic criteria including AIC \citep{akaike-1974-anew}, BIC \citep{schwarz-1978-estimating}, and MDL \citep{rissanen-1978-modeling, grunwald-2007-minimum}, alongside entropy-based filtering methods. To improve estimation under limited data, we augment adaptation corpora with synthetic text generated from the original data, improving unigram and bigram coverage while reducing sparsity.

Our contributions are as follows:
  (a) we formulate BPE merge selection as a model selection problem rather than relying on predefined values;
  (b) we evaluate several statistical and information-theoretic model selection criteria;
  (c) we propose synthetic text augmentation to improve the estimation of token distributions, particularly in low-resource settings;
  (d) we provide empirical analyses showing how different selection criteria affect merge inventories, compression efficiency and downstream performance.

\section{Related Work} \label{sec:related}

\paragraph{Subword Tokenization and Vocabulary Design.}
Subword tokenization methods such as Byte-Pair Encoding (BPE) \citep{gage-1994-anewalgorithm, sennrich-etal-2016-neural}, WordPiece \citep{wu-2016-google}, and Unigram Language Models \citep{kudo-2018-subword} are widely used in modern language models. These methods construct vocabularies by iteratively merging or selecting subword units based on corpus statistics. In practice, the number of merges or vocabulary size is typically fixed heuristically, despite its strong impact on sequence length, compression efficiency and downstream performance \citep{dagan-etal-2024-getting}.

\paragraph{Tokenizer Adaptation and Efficiency.}
Prior work has studied tokenizer adaptation in multilingual and domain-transfer settings, where vocabularies are modified or extended to improve coverage for new languages or domains \citep{sachidananda-etal-2021-efficient, minixhofer-etal-2022-wechsel,remy-etal-2024-transtokenization}. Other studies have analyzed the tradeoff between vocabulary size, sequence length, and computational efficiency, showing that tokenizer design can significantly affect both training and inference cost \citep{sennrich-etal-2016-neural,kudo-2018-subword, zouhar-etal-2023-tokenization, dagan-etal-2024-getting}. In contrast to these approaches, we focus on selecting a compact subset of merges from an existing tokenizer rather than expanding or retraining the vocabulary. Closer to our work, \citep{meister-etal-2026-langmap} trains language-adapted unigram models that all share the same underlying unit set, with which it is then possible to jointly perform tokenization and language identification.

\paragraph{Information-Theoretic Model Selection.}
Information-theoretic criteria such as AIC, BIC \citep{akaike-1974-anew, schwarz-1978-estimating}, and Minimum Description Length (MDL) \citep{rissanen-1978-modeling, grunwald-2007-minimum} provide principled ways to balance model fit and complexity. These criteria have been widely used in statistical modeling and structure learning, but their application to subword tokenization has never, to the best of our knowledge, been considered. We leverage these criteria to formalize the problem of adaptation the vocabulary size to a given task.

\paragraph{Synthetic Data for Distribution Estimation.}
Data generation has been used to improve data coverage in low-resource and distributionally skewed settings. Recent work shows that large language models can generate synthetic corpora that improve robustness in downstream tasks \citep{anikina-etal-2025-rigorous,luo-etal-2025-survey}. In this work, we recourse to synthetic data to improve the estimation of token statistics for merge selection, particularly in low-resource and morphologically rich languages where sparsity affects frequency estimation.

\section{Method \label{sec:method}}

We address the problem of selecting an appropriate subset of BPE merge operations from an initial merge list.\done\todo{This is the most important part - emphasize} Given a training corpus $\corpus$, standard BPE recursively applies a sequence of $N$ merges, where $N$ is a predefined parameter which controls the final vocabulary size. This vocabulary is then used uniformly for all tasks, domains, and language. Our goal is to \emph{select} an optimal subset of merges from this basic vocabulary, based on the statistics of some data $\corpus$, which can be the corpus used to train $T$, or, as we consider in our experiments, an adaptation corpus used to trim the vocabulary.

\subsection{Background} \label{ssec:background}

A tokenizer is a function $T$ mapping sequences of words\footnote{More precisely, of units identified by a pre-tokenizer, segmenting the raw input based on a rational expression.} into a sequence of indexes. Tokenization amounts to segmenting words into subword units in a finite inventory \vocab. Modern LLMs used trained tokenizers, which identify these units either bottom-up, growing a basic unit set via a recursive sequence of merges as in the BPE algorithm of \citep{gage-1994-anewalgorithm}, or top-down, iteratively pruning the least useful units in a very large unit set, as in the Unigram algorithm \citep{kudo-richardson-2018-sentencepiece}.

In this work, we focus on BPE tokenizers. This is because training a BPE tokenizer with a target vocabulary size of $N$ yields a sequence of tokenizers $T_1 \dots T_N$ of increasing vocabulary sizes $U_1 \dots U_N$ , with $T_{j}$ and $T_{j+1}$ only differing by one vocabulary unit. This basic property is key for our selection algorithm, as we explain below.

\subsection{Merge Selection as Model Selection}\label{ssec:model_selection}

Given a training corpus $\corpus$,
we identify a tokenizer $T$ comprising $U$ units with a unigram model $P(u, \vtheta^{T})$ trained on $T(\corpus)$; parameter $\theta_u^T$ corresponds to the unigram probability of token $u$, $l(u)$ is its length. $\ell(\mathcal{D}; \vtheta^{T})$ denotes the log-likelihood of the corpus under model $P(u; \vtheta^{T})$. We also define $f(u; T(\corpus))$ the frequency of $u$ in $T(\corpus)$, $f_{\max}(T(\corpus)) \defun \max_u f(u; T(\corpus))$ the maximum frequency and $S(T(\corpus)) = \sum_u f(u; T(\corpus))$ the corpus size.\footnote{We drop the dependency in $T(\corpus)$ whenever possible to avoid notational clutter.}

We consider model selection criteria from the literature, all combining a log-likelihood term and a model complexity term:

\begin{itemize}
    \item \textbf{AIC:} $-2 \ell(\corpus; \vtheta^T) + (U - 1)$ is the \emph{Akaike Information Criterium} \citep{akaike-1974-anew}, which trades off perplexity and vocabulary size. For AIC, the model complexity term assigns a uniform cost of 1 to all parameters;\footnote{As the unigram probabilities sum to $1$, the model contains only $U-1$ free parameters.}
    \item \textbf{BIC:} $-2 \ell(\corpus; \vtheta^T) + 2 (U - 1) \log S$ is the \emph{Bayesian Information Criterium} \cite{schwarz-1978-estimating}, which further takes the corpus size $S$ into consideration: larger training data enables to estimate larger models;
    \item \textbf{MDL-2-part:} $- \ell(\corpus; \vtheta^T) + (U - 1) \log f_{\max}$ is a basic version of the \emph{Minimum Description Length} (MDL) criteria \citep{rissanen-1978-modeling}. $\log f_{\max}$ corresponds to a coding cost that needs to be ``paid'' for each parameter $\theta_u^T$: assuming maximum likelihood (ML) estimates, $\theta_u^T$ can be identified with count $f(u)$, the encoding of which requires $\log f_{\max}$ bits;
    \item \textbf{MDL-COMP:} $- \ell(\mathcal{D}, \theta) + 0.5 (U - 1) \log S$. This is the revised version of MDL \citep{grunwald-2007-minimum}, with a slightly different formulation of the complexity term. The refined MDL criterion replaces the parameter encoding cost of two-part MDL with an approximation of the model's stochastic complexity, which measures the flexibility of the model class. As a result, the complexity penalty scales with both the number of parameters and the corpus size. \done\todo{Explain ?}
\end{itemize}

In addition, we consider the entropy-based criterion of \citep{xu-etal-2021-vocabulary}:
\begin{itemize}
    \item \textbf{VOLT:}
    \[
    \frac{\sum_u l_u}{U \sum_u f(u)} \times
    \sum_u f(u)\log\frac{f(u)}{\sum_u f(u)}.
    \]
\end{itemize}
The similarity between VOLT and other model selection criteria stems from the fact that the second term in VOLT is equivalent to the log-likelihood of the unigram model (assuming standard ML estimates: $\theta_u^T = \frac{f(u)}{\sum_u f(u)}$). The first term corresponds to the average token length divided by the total number of tokens. The numerator increases as more units are merged, while the denominator decreases, since larger vocabularies imply coarser tokenizations. VOLT can thus also be viewed as performing a (multiplicative) tradeoff between model complexity (expressed by the first term) and goodness of fit (the second term).  

\subsection{Merge Filtering Procedure}

We interpret the BPE merging procedure as a sequence of local model updates, where each candidate merge introduces a new unit and modifies the corpus tokenization. Each merge therefore affects both the \textbf{fit to the data} (via changes in token frequencies and likelihood) and the \textbf{model complexity} (chiefly, via the vocabulary size).

\paragraph{An Algorithm for Incremental Merge Evaluation.}
We assume a existing BPE tokenizer, associated with a predefined merge list and decide which merge to keep or drop based on the following iterative procedure. At step $j$, $\corpus{}$ is tokenized with tokenizer $T_{j-1}$, and we need to evaluate merge $\mu_j$ (creating unit $u_j$). For this, we compute:
\[
\Delta \mathcal{J}(\mu_j) \;=\; \mathcal{J}(\corpus; T_j) \;-\; \mathcal{J}(\corpus; T_{j-1})
\]
with $\mathcal{J}(\corpus; T_J)$ a model selection criterion for tokenizer $T_j$. The merge is accepted if and only if:
\[
\Delta \mathcal{J}(\mu_j) < 0,
\]
i.e., the improvement in data fit outweighs the increase in model complexity. Else, the merge is rejected and the next merge is evaluated. Note that rejecting a merge $\mu_j$ will automatically prevent any unit derived from $u_j$ to be formed at a later stage, as $u_j$ will have a null frequency. The procedure terminates when all merges have been evaluated. It is routine to check that the resulting merge list is valid \citep{berglund-vandermerwe-2023-formalizing} as the validity property is an invariant property wich is maintained at all steps of the procedure.

\paragraph{Merge Refinement}
The procedure sketched above is greedy, as it selects or rejects merges based on a local evaluation of their utility. Like for standard BPE, this process may create to \emph{junk} units \citep{bostrom-durrett-2020-byte}, which, once created, may participate in further merges that decrease their utility. We thus follow \citep{lian-2025-scaffold} and adopt their refinement strategy, which performs a second filtering pass, unmerging, as it were, any unit whose global utility is too small.

\paragraph{Merge Refinement with Scaffolding.}
We now describe the full procedure used to select and refine BPE merges. The algorithm contain two phases: (i) a greedy filtering stage that selects merges based on the offset between goodness-of-fit and complexity,
and (ii) a scaffolding refinement stage that replaces low-utility merges with higher-frequency candidates based on corpus statistics.
The full procedure is formalized in Algorithm~\ref{alg:comp-scaffold-nml} (appendix~\ref{alg:comp-scaffold-nml}).\done\todo{Check notations}

\subsection{Synthetic Data Augmentation}
Our merge selection procedure applies irrespective of the nature of $\corpus$, which can either be the original tokenizer training data, or a new corpus on which the model and its tokenizer need to be adapted. In the latter case, estimating merge utility from limited data can yield unmotivated choices, particularly for low-frequency or unseen bigrams.
In BPE-based adaptation algorithms, merges corresponding to zero-frequency bigrams are never considered, even if they are plausible under the underlying data distribution.

To address this limitation, we augment the adaptation corpus $\mathcal{A}$ with synthetic text generated from a pretrained LLM
. The goal is to better approximate the bigram distribution in the adaptation domain and to recover potentially useful merges that would otherwise remain undetected.

\paragraph{Generation Procedure}

We construct a synthetic corpus adaptation $\mathcal{G}$ by sampling prefixes $p \in \mathcal{A}$ and generating continuations conditioned on these prefixes. Each prefix serves as a prompt to the LLM, yielding texts that preserve local syntactic and semantic structure, while also reflecting the peculiarities of the adaptation corpus.
In the sequel, we note the synthetic corpus derived from $\mathcal{A}$ as $\mathcal{A}_G$.

\begin{figure}[t]
\centering
\includegraphics[width=0.7\columnwidth]{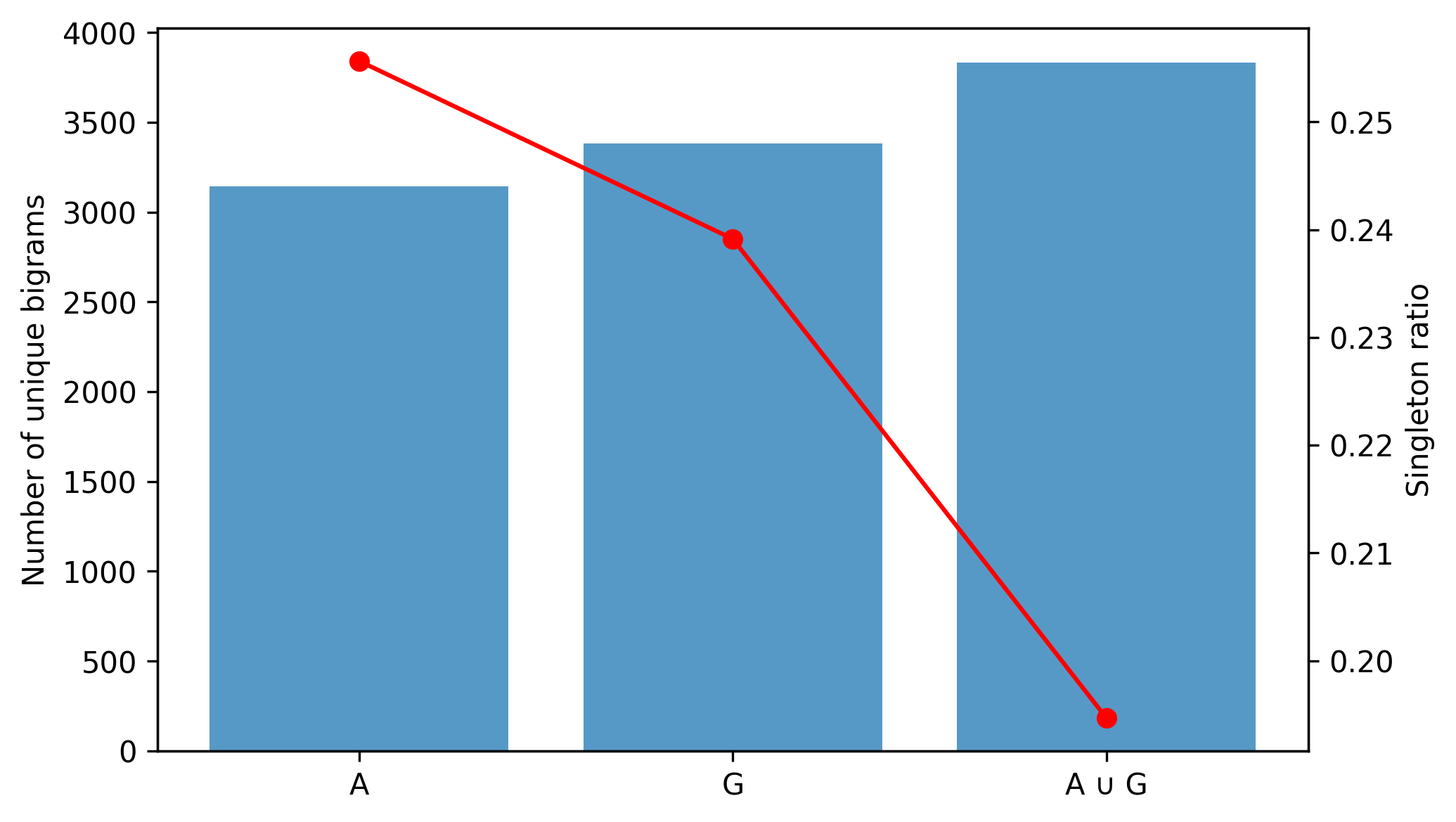}
\caption{Bigram distribution and singleton ratio computed for the EMEA English corpus. We compare using 10k original data ($\mathcal{A}$), 10k generated data ($\mathcal{G}$), and their combination ($\mathcal{A} \cup \mathcal{G}$).}
\label{fig:bigram-distribution}
\end{figure}

By aggregating generated continuations across diverse prefixes, the synthetic corpus approximates the underlying token distribution and increase support for rare or even unseen bigrams. Figure~\ref{fig:bigram-distribution} illustrates this effect via increased bigram coverage and reduced singleton ratio, indicating lower distributional sparsity.

\paragraph{Interpretation}

From a modeling perspective, synthetic data enables to smooth empirical counts and yields more robust estimation of merge utility. This is particularly important for statistical criteria such as AIC, BIC, and MDL, which depend on accurate likelihood estimates.

In our experiments, we evaluate merge selection under three conditions: \textbf{orig} (original corpus only), \textbf{gen} (synthetic corpus only) and \textbf{orig+gen} (combined corpus).
This setup allows us to isolate the contribution of synthetic data and assess their impact on merge selection and downstream tasks.

\section{Experimental Setup} \label{sec:experiments}

\subsection{Corpora}

We evaluate merge selection across multiple corpora covering both domain-specific and multilingual settings.

\paragraph{Domain-specific corpora}
We use the EMEA corpus for English and French,\footnote{EMEA is the European Medicines Agency from the Opus collection \url{https://opus.eu}, \citep{tiedemann-2012-parallel}.} and the LER corpus for German.\footnote{LER is the German Legal Entity Recognition corpus of \citep{leitner-etal-2020-dataset}. See: \url{https://huggingface.co/datasets/german-tokenizer-benchmark/german-ler}.} These datasets provide controlled, domain-specific text suitable for analyzing tokenization under limited variability.

\paragraph{Multilingual corpora}

To assess robustness across languages, we again use subsets of the OPUS collection for English, Sinhala, and Swahili, representing varying levels of resource availability and morphological complexity. For each corpus, we construct training subsets of size 5k, 10k, 20k, and 50k sentences to study the effect of data scale on merge selection.

\subsection{Synthetic Data Generation}

Synthetic text is generated using LLaMA-3.1-8B-instruct\done\todo{instruct?} with nucleus sampling (top-$p=0.9$) and temperature $0.7$. Each prompt consists of a 12-word prefix sampled from the original corpus, followed by an instruction to continue generation until sentence completion. For each prefix, we sample $m$ independent continuations with a maximum length of 64 tokens each.



\subsection{Tokenizer Construction}

All tokenizers are derived from a shared initial BPE merge list. Each selection strategy (AIC, BIC, MDL variants, VOLT) produces a filtered subset of merges.
To ensure comparability, all methods and baselines only differ in their selection criterion, and  operate on the same initial merge inventory.
\done\todo{Also baselines}

\subsection{Evaluation: Tasks and Metrics \label{ssec:tasks}}

\paragraph{Compression Utility} We assess adapted tokenizer based on their ability to compress the adpatation corpus, i.e. to reduce the initial byte based representation (or character-based representation) into a shorter list of tokens. This can be measured in various ways (e.g., tokenizer fertility); in this work we rely on the \emph{compression utility}, which measure the difference (in number of units) between the byte-based and token-based representations as a ratio of the original number of bytes or characters.\done\todo{Or is it the ratio?}
Additionally, we also evaluate the adapted tokenizers on downstream tasks that probe both understanding and generation.

\paragraph{Reading Comprehension}
We evaluate multilingual understanding using the Belebele benchmark \citep{bandarkar-etal-2024-belebele}. Performance in multiple choice question is measured using accuracy. For each example, the model is prompted with the passage, question and four candidate answers, and instructed to output one of the answer labels \texttt{A}, \texttt{B}, \texttt{C}, or \texttt{D}.
The predicted answer is obtained by extracting the generated choice label, then compared against the gold answer to compute accuracy. \done\todo{How are predictions recorded?}

\paragraph{Machine Translation}
To assess generation, we use Machine Translation (MT). As MT requires (at least) two languages, we build the adaptation corpus (and the corresponding synthetic data) using a balanced subset of sentences sampled from the source and target corpus. We use Flores-200 for multilingual translation \citep{costajussa-etal-2024-scaling}. Translation quality is measured in BLEU \citep{papineni-etal-2002-bleu} implemented with \texttt{sacreBLEU} \citep{post-2018-call}.\footnote{Signature: \texttt{nrefs:1|case:mixed|eff:no|tok:13a| \linebreak smooth:exp|version:2.5.1}} We use greedy decoding with a maximum generation limit of 128~tokens for all models.

\subsection{Baselines}
We compare our new merge selection strategies to Adapt-BPE \citep{pilana-liyanage-yvon-2026-adaptbpe}, which also adapts a tokenizer to a target corpus, using a comparable merge budget. For all pruning-based approaches (Adapt-BPE, AIC, BIC, MDL-2-part, MDL-COMP, and VOLT), we use a fixed budget of 15k merges to ensure a fair comparison. 
\done \todo{What are the baselines - fixed 15? AdaptBPE}

\subsection{Experimental Protocol}

For each corpus, dataset size, and data condition (\textbf{orig}, \textbf{gen}, \textbf{orig+gen}), our experiments rely on the following steps: 
  (a) construct tokenizers using various selection strategies;
  (b) apply the resulting tokenizers to downstream tasks;
  (c) evaluate compression and task performance.
All experiments use the same underlying model and inference settings, ensuring that differences in performance arise solely from the choice of tokenizer.

\section{Results} \label{sec:results}
\done\todo{Improve presentation of results}
We present results across three evaluation dimensions: compression utility (CU), reading comprehension (RC), and machine translation (MT). 

\subsection{Compression Utility}
\done\todo{What is the initial tokenizer? What is its size?}\done\todo{The caption says gains - are these number relative ? to what?}\done\todo{Why is not VOLT the best option as it has a larger unit set?}
 \begin{figure*}[t]
     \centering
     \includegraphics[width=\linewidth]{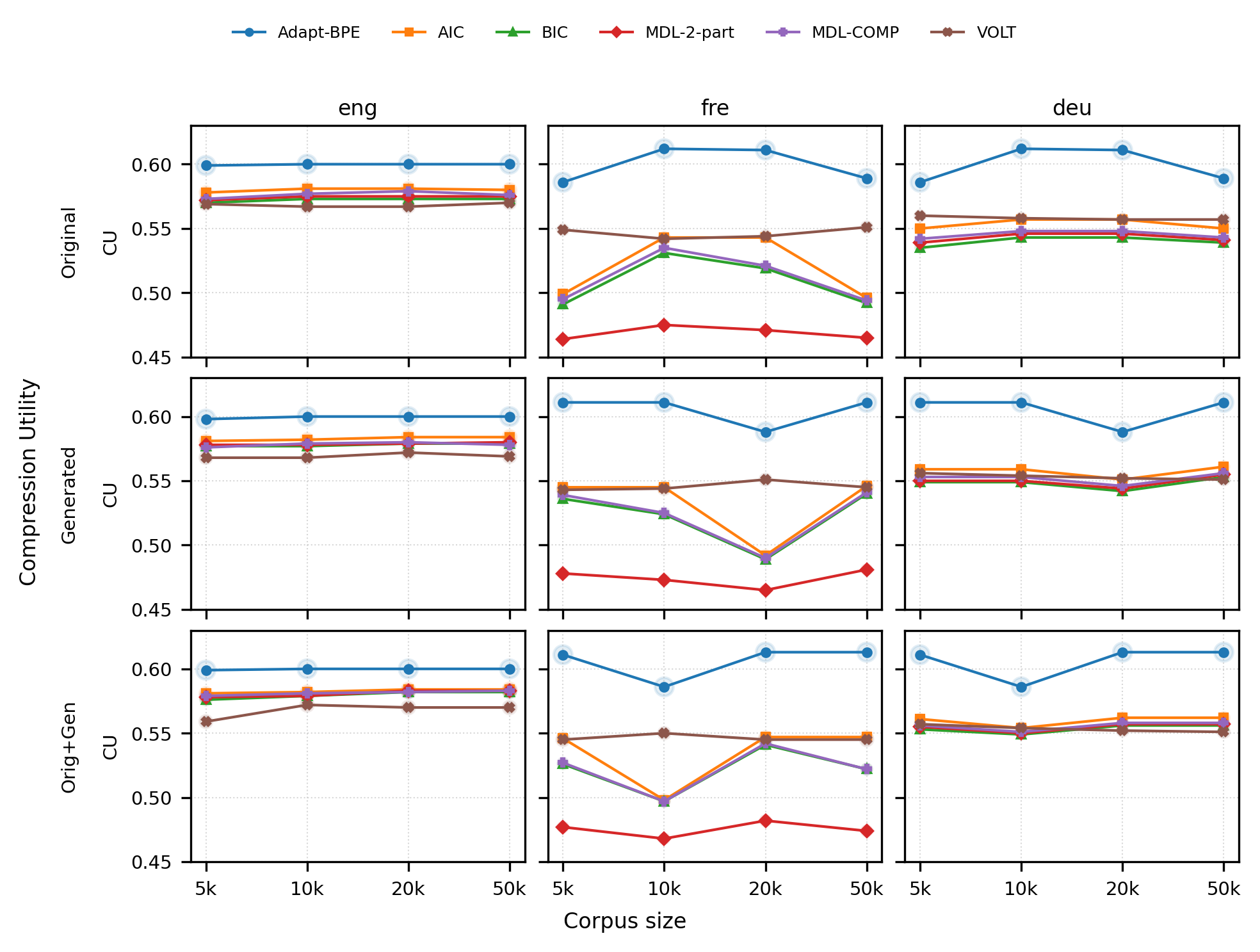}
     \caption{CU values for each strategy, corpus types and sizes in English, French and German.}
     \label{fig:HM-CU}
\end{figure*}

\done \todo{Is this the number of units filtered or kept?}\done\todo{I am not a fan of Heatmaps here}
Figure~\ref{fig:HM-CU}\done\todo{The language names are not right in the figure} reports CU for English, French, and German corpora of varying sizes. For these experiments, we  used the Llama-3 tokenizer (128k) as the initial tokenizer. For all languages, the 15k-baseline consistently achieves the highest compression utility, reflecting the advantage of a large predefined merge budget. However, all selection-based methods operate with much smaller vocabularies - only retaining between 3 times and 10 times less units (full numbers for the 50k adaptation corpus are in Table~\ref{tab:Belebele-50k}). Although \textbf{VOLT} generally retains a larger number of merges than the information-theoretic methods, it does not consistently achieve the highest CU. This suggests that the effectiveness of merge selection depends not only on the number of retained merges but also on selecting the most informative ones, where MDL-COMP and AIC provide a better tradeoff between vocabulary size and compression.\todo{Perhaps having a graph equivalent to Fig. 2 in appendix with vocab sizes would help.}

Among the adaptive selection methods, \textbf{VOLT} and \textbf{AIC} consistently outperform \textbf{MDL-*} and \textbf{BIC}, yielding larger unit sets, especially for the original corpus. In particular, \textbf{BIC} tends to underfit, selecting smaller vocabularies that lead to lower CU than the other methods. We also observe variance across languages, owing to the initial vocabulary content for those languages.
\done\todo{where is VOLT?}




\subsection{Reading Comprehension}

Table~\ref{tab:Belebele-50k}\done\todo{One digit accuracy is enough} reports accuracy on the Belebele benchmark across multiple languages.
\done\todo{With or without BPE Scaffolding?}
\done\todo{Why 30K for adapt BPE? The text says 15k?}
\begin{table*}[htbp]
\centering
\resizebox{\textwidth}{!}{%
\begin{tabular}{lccccccccccccccc}
\toprule
\multirow{2}{*}{\textbf{Strategy}}
& \multicolumn{5}{c}{\textbf{orig}} 
& \multicolumn{5}{c}{\textbf{gen}} 
& \multicolumn{5}{c}{\textbf{orig+gen}} \\
\cmidrule(lr){2-6} \cmidrule(lr){7-11} \cmidrule(lr){12-16}
  & eng & fre & deu & swh & sin 
  & eng & fre & deu & swh & sin  
  & eng & fre & deu & swh & sin  \\
 \midrule

 Full (128k)
 & 81.7 & 74.1 & 75.0 & 49.7 & 48.3 
 & 81.7 & 74.1 & 75.0 & 49.7 & 48.3 
 & 81.7 & 74.1 & 75.0 & 49.7 & 48.3 \\

 Adapt-BPE (15k)
 & 70.3 & 68.3 & 70.0 & 45.8 & 48.3
 & 68.2 & 66.7 & 67.8 & 43.9 & 47.7 
 & 71.6 & 69.5 & 70.0 & 45.8 & 48.3\\

AIC 
& 60.0 (5.9k) & 58.9 (2.5k) & 60.0 (3.1k) & 42.2 (2.4k) & 46.5 (1.7k)
& 58.0 (6.6k) & 57.7 (3.4k) & 59.9 (2.6k) & 40.4 (2.6k) & 46.0 (1.6k)
& 60.1 (6.6k) & 58.9 (3.8k) & 60.1 (3.2k) & 42.3 (2.7k) & 47.0 (1.7k)\\

\quad + Scaffolding
& 60.3 (5.9k) & 60.4 (2.5k) & \textbf{62.1} (3.1k) & 43.7 (2.4k) & 47.9 (1.7k)
& 59.9 (6.6k) & 58.3 (3.4k) & \textbf{60.1} (2.6k) & 42.2 (2.6k) & 46.5 (1.6k)
& 61.6 (6.6k) & 60.9 (3.8k) & \textbf{62.1} (3.2k) & 43.4 (2.7k) & 48.0 (1.7k)\\

BIC 
& 56.2 (4.1k) & 55.5 (1.4k) & 57.7 (2.3k) & 40.1 (1.9k) & 45.6 (1.4k)
& 55.9 (3.7k) & 55.2 (2.0k) & 56.8 (1.6k) & 39.9 (2.0k) & 44.2 (1.4k)
& 56.5 (5.0k) & 57.3 (2.4k) & 58.8 (2.4k) & 42.1 (2.1k) & 46.6 (1.5k)\\

\quad + Scaffolding
& 57.1 (4.1k) & 56.9 (1.4k) & 58.9 (2.3k) & 42.1 (1.9k) & 46.9 (1.4k)
& 57.0 (3.7k) & 56.0 (2.0k) & 57.2 (1.6k) & 41.1 (2.0k) & 46.2 (1.4k)
& 57.5 (5.0k) & 57.1 (2.4k) & 59.3 (2.4k) & 43.1 (2.1k) & 47.3 (1.5k)\\

MDL-2-part 
& 57.7 (3.1k) & 57.0 (1.5k) & 58.8 (2.4k) & 41.1 (2.1k) & 46.8 (1.6k)
& 56.7 (4.0k) & 56.6 (2.1k) & 58.1 (1.7k) & 40.1 (2.2k) & 46.0 (1.7k)
& 58.0 (5.3k) & 57.7 (2.5k) & 60.0 (2.5k) & 42.3 (2.3k) & 46.8 (1.7k)\\

\quad + Scaffolding
& 58.1 (3.1k) & 58.1 (1.5k) & 59.1 (2.4k) & 42.3 (2.1k) & 47.8 (1.6k)
& 57.1 (4.0k) & 57.2 (2.1k) & 59.0 (1.7k) & 42.1 (2.2k) & 46.5 (1.7k)
& 60.1 (5.3k) & 60.1 (2.5k) & 60.1 (2.5k) & 43.0 (2.3k) & 48.0 (1.7k)\\

MDL-COMP 
& 58.8 (3.5k) & 57.7 (1.7k) & 57.8 (2.6k) & 42.2 (2.3k) & 47.9 (1.9k)
& 58.0 (4.4k) & 57.4 (2.8k) & 57.7 (1.9k) & 42.5 (2.2k) & 47.0 (1.9k)
& 59.0 (5.5k) & 58.8 (2.8k) & 58.0 (2.7k) & 43.2 (2.3k) & 48.0 (1.9k)\\

\quad + Scaffolding
& 60.0 (3.5k) & 58.1 (1.7k) & 58.9 (2.6k) & 43.8 (2.3k) & 48.1 (1.9k)
& 58.8 (4.4k) & 58.0 (2.8k) & 58.7 (1.9k) & 43.0 (2.2k) & 48.0 (1.9k)
& 60.5 (5.5k) & 60.0 (2.8k) & 58.9 (2.7k) & 44.2 (2.3k) & 48.2 (1.9k)\\

VOLT
& 59.3 (4.7k) & 60.0 (3.4k) & 59.1 (3.5k) & 45.2 (2.4k) & 47.6 (2.1k)
& 59.0 (5.2k) & 59.6 (4.6k) & 59.0 (3.4k) & 45.6 (2.5k) & 47.5 (2.3k)
& 60.0 (5.9k) & 60.1 (5.4k) & 60.2 (3.9k) & 45.9 (2.7k) & 48.2 (2.3k)\\

\quad + Scaffolding
& \textbf{61.0} (4.7k) & \textbf{61.0} (3.4k) & 61.0 (3.5k) & \textbf{47.2} (2.4k) & \textbf{48.2} (2.1k)
& \textbf{60.9} (5.2k) & \textbf{60.2} (4.6k) & 59.8 (3.4k) & \textbf{46.3} (2.5k) & \textbf{48.0} (2.3k)
& \textbf{62.1} (5.9k) & \textbf{62.9} (5.4k) & 61.2 (3.9k) & \textbf{47.9} (2.7k) & \textbf{48.3} (2.3k)\\

\bottomrule
\end{tabular}
}
\caption{RC results (Accuracy). For AIC, BIC, MDL and VOLT, merge lists are filtered using a corpus of 50k lines.}
\label{tab:Belebele-50k}
\end{table*}

In these experiments, the full tokenizer achieves the highest accuracy across the board, followed by the fixed-15k baseline. Among the selection strategies, VOLT and MDL-based methods (particularly MDL-COMP) consistently outperform AIC and BIC.

The performance gap between large and reduced vocabularies is significant for high-resource languages. For the two LR languages (sin and swh), for which the baseline performance is low, it is possible to perform a very agressive vocabulary pruning with little impact on performance, sometimes even outperforming the baseline. This highlights the unfairness of the initial tokenizer, which allocates a lots of units to improve the tokenization of HR languages, while the LR languages are comparatively less well served.

We further observe that BPE scaffolding consistently improves performance across nearly all filtering strategies and data settings. The gains are particularly noticeable for VOLT and MDL-based methods. 

Finally, synthetic data again provides modest but consistent improvements when combined with actual data.

\subsection{Machine Translation} \label{ssec:machinetranslation}

Table~\ref{tab:MT-sin-eng}\done\todo{report only one digit of precision to increase the font size} reports BLEU scores for Sinhala-English translation on Flores-200.\footnote{French-English results are in Appendix~\ref{app:fre-eng-results}.} The full tokenizer and the 15k Adapt-BPE baseline\done\todo{30k ?} achieve the strongest overall performance, confirming the advantage of larger merge inventories. Among the selection-based approaches, however, \textbf{VOLT} and \textbf{MDL-COMP} consistently outperform AIC and BIC, despite relying on substantially smaller vocabularies.
Here again, scaffolding consistently improves BLEU scores across nearly all selection strategies. The gains are particularly strong for VOLT and MDL-based methods, indicating that greedy merge filtering alone may discard useful subword units that can later be recovered through refinement.

Synthetic data also provides clear benefits. Across most settings, the \textbf{orig+gen condition achieves the best performance}, especially for smaller corpora, suggesting that generated text improves estimation of low-frequency and previously unseen token patterns. These improvements are particularly important in the Sinhala-English setting, where limited adaptation data makes merge estimation more sensitive to sparsity.

Interestingly, larger vocabularies do not always yield better translation quality. For example, AIC often selects more merges than MDL-COMP while achieving lower BLEU scores.

\begin{table*}[h!]
\centering
\resizebox{\textwidth}{!}{%
\begin{tabular}{lcccccccccccc}
\toprule
\multirow{2}{*}{\textbf{Strategy}}
& \multicolumn{4}{c}{\textbf{orig}} 
& \multicolumn{4}{c}{\textbf{gen}} 
& \multicolumn{4}{c}{\textbf{orig+gen}} \\
\cmidrule(lr){2-5} \cmidrule(lr){6-9} \cmidrule(lr){10-13}
 & 5k+5k & 10k+10k & 20k+20k & 50k+50k 
 & 5k+5k & 10k+10k & 20k+20k & 50k+50k  
 & 5k+5k & 10k+10k & 20k+20k & 50k+50k  \\
\midrule

Adapt-BPE
& 22.6 (30k) & 22.6 (30k) & 22.7 (30k) & 22.7 (30k)
& 21.1 (30k) & 21.1 (30k) & 21.3 (30k) & 21.4 (30k)
& 23.6 (30k) & 23.6 (30k) & 23.7 (30k) & 23.8 (30k)\\

AIC 
& 21.3 (4.4k) & 21.3 (5.0k) & 21.3 (5.5k) & 21.4 (6.0k)
& 21.0 (4.7k) & 21.0 (5.3k) & 21.1 (6.0k) & 21.1 (6.6k)
& 21.3 (4.7k) & 21.3 (5.3k) & 21.3 (6.0k) &  21.4 (6.6k)\\

\quad + Scaffolding
& \textbf{21.4} (4.4k) & 21.5 (5.0k) & 21.5 (5.5k) & 21.6 (6.0k)
& \textbf{22.0} (4.7k) & \textbf{22.1} (5.3k) & \textbf{22.2} (6.0k) & \textbf{22.2} (6.6k)
& \textbf{22.4} (4.7k) & 22.4 (5.3k) & 22.4 (6.0k) &  22.5 (6.6k)\\

BIC
& 17.1 (2.3k) & 17.2 (2.9k) & 17.2 (3.5k) & 17.7 (4.1k)
& 16.9 (2.4k) & 16.9 (3.0k) & 16.9 (3.7k) & 17.0 (3.7k)
& 17.2 (3.1k) & 17.2 (3.7k) & 17.2 (4.5k) & 17.7 (5.0k)\\

\quad + Scaffolding
& 18.2 (2.3k) & 18.2 (2.9k) & 18.2 (3.5k) & 18.7 (4.1k)
& 18.0 (2.4k) & 18.0 (3.0k) & 18.0 (3.7k) & 18.1 (3.7k)
& 18.3 (3.1k) & 18.3 (3.7k) & 18.3 (4.5k) & 18.4 (5.0k)\\

MDL-2-part 
& 19.3 (2.6k) & 19.5 (2.7k) & 19.8 (2.7k) & 19.8 (3.1k)
& 18.0 (3.3k) & 18.0 (3.3k) & 18.1 (3.7k) & 18.1 (4.0k)
& 20.0 (3.8k) & 20.2 (4.4k) & 20.3 (5.3k) & 20.7 (5.3k)\\

\quad + Scaffolding
& 20.5 (2.6k) & 20.5 (2.7k) & 20.8 (2.7k) & 20.9 (3.1k)
& 21.0 (3.3k) & 20.1 (3.3k) & 20.2 (3.7k) & 20.2 (4.0k)
& 21.1 (3.8k) & 21.5 (4.4k) & 21.7 (5.3k) & 21.8 (5.3k)\\

MDL-COMP 
& 20.1 (3.0k) & 20.3 (3.1k) & 20.3 (3.7k) & 20.4 (3.5k)
& 20.0 (3.7k) & 20.0 (3.7k) & 20.1 (4.4k) & 20.1 (4.4k)
& 20.3 (4.0k) & 20.4 (4.4k) & 21.1 (5.1k) & 21.2 (5.5k)\\

\quad + Scaffolding
& 21.3 (3.0k) & 21.5 (3.1k) & 21.5 (3.7k) & 21.5 (3.5k)
& 21.0 (3.7k) & 21.0 (3.7k) & 21.2 (4.4k) & 21.2 (4.4k)
& 21.5 (4.0k) & 21.5 (4.4k) & 21.5 (5.1k) & 22.2 (5.5k)\\

VOLT
& 20.32 (4.2k) & 20.3 (4.7k) & 20.4 (4.4k) & 20.4 (4.7k)
& 19.8 (4.6k) & 19.9 (4.6k) & 20.0 (5.3k) & 20.3 (5.2k)
& 20.5 (5.1k) & 21.9 (5.9k) & 21.9 (5.9k) & 22.3 (5.9k) \\

\quad + Scaffolding
& 22.3 (4.2k) & \textbf{22.4} (4.7k)  & \textbf{22.4} (4.4k) & \textbf{22.4} (4.7k) 
& 21.3 (4.6k) & 21.5 (4.6k) & 21.5 (5.3k) & 21.5 (5.2k)
& 22.1 (5.1k) & \textbf{22.9} (5.9k) & \textbf{23.2} (5.9k) & \textbf{23.7} (5.9k) \\

\bottomrule
\end{tabular}
}
\caption{MT results (BLEU) for sin-eng Flores-200 data. Using the full tokenizer yields a BLEU score of 23.2.}
\label{tab:MT-sin-eng}
\end{table*}

\subsection{Increasing Synthetic Data}

\begin{table*}[h!]
\centering
\small
\resizebox{\textwidth}{!}{%
\begin{tabular}{l ccc cc ccc ccc cc ccc}
\toprule
\textbf{Method} 
& \multicolumn{8}{c}{\textbf{100k}} 
& \multicolumn{8}{c}{\textbf{200k}} \\
\cmidrule(lr){2-9} \cmidrule(lr){10-17}

& \multicolumn{3}{c}{$\Delta$CU}
& \multicolumn{2}{c}{$\Delta$BLEU} 
& \multicolumn{3}{c}{$\Delta$Acc}
& \multicolumn{3}{c}{$\Delta$CU}
& \multicolumn{2}{c}{$\Delta$BLEU} 
& \multicolumn{3}{c}{$\Delta$Acc} \\
\cmidrule(lr){2-4} \cmidrule(lr){5-6} \cmidrule(lr){7-9}
\cmidrule(lr){10-12} \cmidrule(lr){13-14} \cmidrule(lr){15-17}

& eng & fre & sin
& fre-eng & sin-eng 
& eng & fre & sin
& eng & fre & sin
& fre-eng & sin-eng 
& eng & fre & sin \\
\midrule

AIC       
& $+4.55$ & $+5.32$ & $+5.85$
& +4.31 & +1.59 
& +6.29 & +4.11 & +0.90
& $+6.13$ & $+5.99$ & $+6.11$ 
& +5.89 & +1.73  
& +6.98 & +5.21 & +1.05 \\

BIC       
& $+8.53$ & $+10.01$ & $+12.13$
& +3.90 & +1.71 
& +6.14 & +4.80 & +1.11
& $+9.42$ & $+11.08$ & $+14.02$ 
& +4.08 & +1.79 
& +6.29 & +4.93 & +1.65 \\

MDL-2Part      
& $+1.65$ & $+1.83$ & $+1.98$ 
& +4.93 & +1.88 
& +5.91 & +4.61 & +1.34
& $+2.05$ & $+1.95$ & $+2.48$  
& +5.13 & +1.93 
& +6.03 & +4.93 & +1.77 \\

MDL-COMP      
& $+19.53$ & $+21.12$ & $+23.46$ 
& +5.03 & +1.93 
& +4.88 & +4.74 & +0.86
& $+24.66$ & $+22.22$ & $+26.17$ 
& +5.29 & +1.99 
& +5.16 & +5.02 & +0.95 \\

VOLT      
& $+7.21$ & $+8.31$ & $+9.12$ 
& +5.45 & +1.78 
& +5.03 & +4.13 & +0.96
& $+8.24$ & $+11.53$ & $+12.95$
& +5.65 & +1.83 
& +5.14 & +4.56 & +1.04 \\

\bottomrule
\end{tabular}
}
\caption{Performance improvements when increasing synthetic data from 50k to 100k and 200k. We report changes ($\Delta$) in Compression Utility (CU$ \times 10^{-4}$) for English, French, and Sinhala, in BLEU (MT for sin-eng and fra-eng), and Accuracies (RC for fre, eng, sin). All results are without scaffolding.}
\label{tab:scaling_results}
\end{table*}

Table~\ref{tab:scaling_results} summarizes the effect of increasing the synthetic data from 50k to 100k, then 200k samples. We observe consistent improvements across all metrics (CU, MT, RC) as the amount of synthetic data increases. The gains are especially pronounced for MDL-COMP and VOLT, suggesting that these methods benefit more from improved estimation of token distributions. Overall, these results confirm that synthetic data acts as an effective smoothing mechanism, enabling better estimation of low-frequency and unseen token patterns and improving merge selection.

To better understand these improvements, we analyze how synthetic data affects token statistics used for merge evaluation. Table~\ref{tab:bigram_synthetic_effect} reports the effect of augmenting the original corpus with synthetic data for a fixed 20k+20k setting. Here, the coverage, is computed as the fraction of test bigrams observed in the training corpus. We observe a substantial increase in observed bigrams (+32\%) for English.\done\todo{which language?}

\begin{table}[h]
\centering
\small
\begin{tabular}{lcc}
\toprule
\textbf{Dataset} & \textbf{Unique Bigrams} 
& \textbf{Coverage} \\
\midrule
orig (20k) 
& 87194 

& 0.1230 \\

orig + gen (20k + 20k) 
& 115597 
& 0.1373 \\
\bottomrule
\end{tabular}
\caption{Effect of synthetic data augmentation on bigram statistics.
}
\label{tab:bigram_synthetic_effect}
\end{table}

\section{Discussion} \label{sec:discussion}

Our results provide several key insights into the problem of tokenizer capacity selection and the role of statistical criteria in guiding BPE merge selection.

\paragraph{Tokenizer Selection as Model Selection}

Across all experiments, we find that treating merge selection as a model selection problem leads to compact vocabularies that retain strong performance. While the fixed-15k baseline achieves the highest scores due to its large predefined merge budget, selection-based methods are able to identify significantly smaller vocabularies that remain competitive across tasks, at a much reduced computational costs. Large merge budgets are not always necessary, and principled selection criteria can yield efficient tokenizer with adaptive vocabulary sizes.

\paragraph{Effect of Model Selection Criteria}

Different selection criteria exhibit distinct behaviors. BIC consistently selects smaller vocabularies, often leading to underfitting due to its strong complexity penalty. AIC allows larger vocabularies, but does not consistently translate this into improved downstream performance. In contrast, MDL-based methods, particularly MDL-COMP, provide a more balanced tradeoff, adapting effectively to different data sizes. VOLT also performs competitively for all tasks. 


\paragraph{Role of Scaffolding Refinement}

The scaffolding refinement stage consistently improves results across all criteria. This confirms that purely greedy merge selection is unefficient, as early decisions may exclude useful merges. By replacing low-frequency merges with high-frequency candidates, scaffolding introduces a form of global correction, leading to better merge inventories.

\paragraph{Impact of Synthetic Data}

Synthetic data improves merge selection, with consistent gains observed across CU, MT, and RC, particularly when combined with natural data (orig+gen setting). 

This supports the interpretation of synthetic data as a distributional prior that reduces sparsity and provides better estimates of rare or unseen token combinations, which is especially important for statistical criteria relying on accurate likelihood estimation. Increasing the amount of synthetic data further improves performance, indicating that merge selection benefits from increasingly better approximations of the underlying distribution.

\paragraph{Task Sensitivity to Tokenization}

We observe that different tasks exhibit different sensitivities to tokenization. Machine translation is relatively robust to vocabulary reduction, whereas reading comprehension is more sensitive, particularly in a multilingual setting, as token fragmentation can affect the representation of rare words, named entities, and morphologically complex forms required for precise semantic matching. In contrast, translation can compensate for increased fragmentation through contextual generation and compositional decoding.

This suggests that tasks requiring precise semantic matching may benefit from fine-grained tokenization, while generative tasks can tolerate more compact representations. Future work could quantify this sensitivity by relating performance degradation under controlled vocabulary reduction to token-level properties such as fertility, named entity fragmentation, and morphological complexity.

%
\done\todo{How to test this?}






\done\todo{Remove all figures not cited in text}




\done\todo{We need some estimates of the computation gains !!! From 5K to 15k what is the speed increase ? I assume small ? : yes it was around 20 miliseconds}

\section{Conclusion} \label{sec:conclusion}

We studied BPE merge inventory selection from a statistical model selection perspective, framing tokenizer capacity as a tradeoff between data likelihood and model complexity through criteria such as AIC, BIC, MDL, and entropy-based objectives.

Across multiple corpora and languages, these criteria produce compact vocabularies that remain competitive on downstream tasks while requiring substantially fewer merges than fixed-budget tokenizers. MDL-based objectives and VOLT provide the most consistent balance between compression efficiency and task performance.

We further show that synthetic data improves merge selection by providing better token statistics, reducing sparsity and improving coverage of rare or unseen combinations, particularly for low-resource and morphologically rich languages.

Overall, our results demonstrate that tokenizer design benefits from principled statistical selection rather than heuristic merge budgets. Downstream performance does not scale monotonically with vocabulary size, and compact inventories can preserve task performance under constrained computational resources.


\done \todo{Also: better understand the interplay between more units and the tasks}

\section*{Limitations}

Our evaluation relies on unigram model to select an optimal unit set, that we then apply to a BPE vocabulary construction process. The choice to combine two techniques makes it easy to implement an incremental vocabulary growth; it also makes our approach fully with complementary techniques such as BPE-scaffolding. However,  it certainly yields to suboptimal unit sets. An alternative would be to keep BPE as a way to incrementally extend the set of units, and to perform tokenization with the unigram model.

Finally, our evaluation is limited to translation and reading comprehension benchmarks, and future work should consider broader tasks and large-scale language model training settings. In addition, evaluating across a wider variety of pretrained tokenizers, model architectures, and languages, particularly underrepresented and morphologically diverse languages, would provide a more comprehensive understanding of the generality and limitations of adaptive vocabulary refinement.

\done\todo{add: more models and tokenizers, more languages and tasks}


\bibliography{anthology,custom}

@article{costajussa-etal-2024-scaling,
	title = {Scaling neural machine translation to 200 languages},
	volume = {630},
	url = {https://doi.org/10.1038/s41586-024-07335-x},
	doi = {10.1038/s41586-024-07335-x},
	number = {8018},
	journal = {Nature},
	author = {Costa-jussà, Marta R. and Cross, James and Çelebi, Onur and Elbayad, Maha and Heafield, Kenneth and Heffernan, Kevin and Kalbassi, Elahe and Lam, Janice and Licht, Daniel and Maillard, Jean and Sun, Anna and Wang, Skyler and Wenzek, Guillaume and Youngblood, Al and Akula, Bapi and Barrault, Loic and Gonzalez, Gabriel Mejia and Hansanti, Prangthip and Hoffman, John and Jarrett, Semarley and Sadagopan, Kaushik Ram and Rowe, Dirk and Spruit, Shannon and Tran, Chau and Andrews, Pierre and Ayan, Necip Fazil and Bhosale, Shruti and Edunov, Sergey and Fan, Angela and Gao, Cynthia and Goswami, Vedanuj and Guzmán, Francisco and Koehn, Philipp and Mourachko, Alexandre and Ropers, Christophe and Saleem, Safiyyah and Schwenk, Holger and Wang, Jeff and Team, NLLB},
	month = jun,
	year = {2024},
	note = {ISBN: 1476-4687
tex.date-added: 2024-08-21 15:32:57 +0200
tex.date-modified: 2024-08-21 15:32:57 +0200},
	pages = {841--846},
}

@misc{meister-etal-2026-langmap,
	title = {{LangMAP}: {A} {Language}-{Adaptive} {Approach} to {Tokenization}},
	shorttitle = {{LangMAP}},
	url = {http://arxiv.org/abs/2606.23566},
	doi = {10.48550/arXiv.2606.23566},
	urldate = {2026-07-03},
	publisher = {arXiv},
	author = {Meister, Clara and Salhan, Suchir and Szablewski, Andrzej and Lesci, Pietro and Buttery, Paula and Pimentel, Tiago},
	month = jun,
	year = {2026},
	note = {arXiv:2606.23566 [cs.CL]},
}

@inproceedings{dagan-etal-2024-getting,
	address = {Vienna, Austria},
	series = {{ICML}'24},
	title = {Getting the most out of your tokenizer for pre-training and domain adaptation},
	booktitle = {Proceedings of the 41st international conference on machine learning},
	publisher = {JMLR.org},
	author = {Dagan, Gautier and Synnaeve, Gabriel and Rozière, Baptiste},
	year = {2024},
	note = {Number of pages: 22
tex.articleno: 387},
}

@article{berglund-vandermerwe-2023-formalizing,
  title = {Formalizing {{BPE}} Tokenization},
  author = {Berglund, Martin and {van der Merwe}, Brink},
  year = 2023,
  month = sep,
  journal = {Electronic Proceedings in Theoretical Computer Science},
  volume = {388},
  pages = {16--27},
  publisher = {Open Publishing Association},
  issn = {2075-2180},
  doi = {10.4204/eptcs.388.4}
}

@article{gage-1994-anewalgorithm,
	title = {A new algorithm for data compression},
	volume = {12},
	issn = {0898-9788},
	url = {https://www.derczynski.com/papers/archive/BPE_Gage.pdf},
	number = {2},
	journal = {Computer Users Journal},
	author = {Gage, Philip},
	month = feb,
	year = {1994},
	pages = {23--38},
}

@article{brown-2020-language,
  title={Language models are few-shot learners},
  author={Brown, Tom and Mann, Benjamin and Ryder, Nick and Subbiah, Melanie and Kaplan, Jared D and Dhariwal, Prafulla and Neelakantan, Arvind and Shyam, Pranav and Sastry, Girish and Askell, Amanda and Agarwal, Sandhini and Herbert-Voss, Ariel and Krueger, Gretchen and Henighan, Tom and Child, Rewon and Ramesh, Aditya and Ziegler, Daniel M. and Wu, Jeffrey and Winter, Clemens and Hesse, Christopher and Chen, Mark and Sigler, Eric and Litwin, Mateusz and Gray, Scott and Chess, Benjamin and Clark, Jack and Berner, Christopher and McCandlish, Sam and Radford, Alec and Sutskever, Ilya and Amodei, Dario},       journal={Advances in neural information processing systems},
  volume={33},
  pages={1877--1901},
  year={2020},
  url=  {https://proceedings.neurips.cc/paper_files/paper/2020/file/1457c0d6bfcb4967418bfb8ac142f64a-Paper.pdf},
  
}

@misc{touvron-etal-2023-llama,
	title = {{LLaMA}: {Open} and {Efficient} {Foundation} {Language} {Models}},
	author = {Touvron, Hugo and Lavril, Thibaut and Izacard, Gautier and Martinet, Xavier and Lachaux, Marie-Anne and Lacroix, Timothée and Rozière, Baptiste and Goyal, Naman and Hambro, Eric and Azhar, Faisal and Rodriguez, Aurelien and Joulin, Armand and Grave, Edouard and Lample, Guillaume},
      year = {2023},
      eprint = {2302.13971},
      archivePrefix={arXiv},
      primaryClass={cs.CL},
      url={https://arxiv.org/abs/2302.13971}, 
}

@misc{bai-2023-qwen,
      title={Qwen Technical Report}, 
      author={Jinze Bai and Shuai Bai and Yunfei Chu and Zeyu Cui and Kai Dang and Xiaodong Deng and Yang Fan and Wenbin Ge and Yu Han and Fei Huang and Binyuan Hui and Luo Ji and Mei Li and Junyang Lin and Runji Lin and Dayiheng Liu and Gao Liu and Chengqiang Lu and Keming Lu and Jianxin Ma and Rui Men and Xingzhang Ren and Xuancheng Ren and Chuanqi Tan and Sinan Tan and Jianhong Tu and Peng Wang and Shijie Wang and Wei Wang and Shengguang Wu and Benfeng Xu and Jin Xu and An Yang and Hao Yang and Jian Yang and Shusheng Yang and Yang Yao and Bowen Yu and Hongyi Yuan and Zheng Yuan and Jianwei Zhang and Xingxuan Zhang and Yichang Zhang and Zhenru Zhang and Chang Zhou and Jingren Zhou and Xiaohuan Zhou and Tianhang Zhu},
      year={2023},
      eprint={2309.16609},
      archivePrefix={arXiv},
      primaryClass={cs.CL},
      url={https://arxiv.org/abs/2309.16609}, 
}

@article{akaike-1974-anew,
  author={Akaike, H.},
  journal={IEEE Transactions on Automatic Control}, 
  title={A new look at the statistical model identification}, 
  year={1974},
  volume={19},
  number={6},
  pages={716-723},
  doi={10.1109/TAC.1974.1100705},
}

@article{schwarz-1978-estimating,
  title={Estimating the dimension of a model},
  author={Schwarz, Gideon},
  journal={The annals of statistics},
  pages={461--464},
  year={1978},
  publisher={JSTOR},
}

@article{rissanen-1978-modeling,
  title={Modeling by shortest data description},
  author={Rissanen, Jorma},
  journal={Automatica},
  volume={14},
  number={5},
  pages={465--471},
  year={1978},
  publisher={Elsevier},
}

@book{grunwald-2007-minimum,
  title={The minimum description length principle},
  author={Gr{\"u}nwald, Peter D},
  year={2007},
  publisher={MIT press},
}

@misc{wu-2016-google,
      title={Google's Neural Machine Translation System: Bridging the Gap between Human and Machine Translation}, 
      author={Yonghui Wu and Mike Schuster and Zhifeng Chen and Quoc V. Le and Mohammad Norouzi and Wolfgang Macherey and Maxim Krikun and Yuan Cao and Qin Gao and Klaus Macherey and Jeff Klingner and Apurva Shah and Melvin Johnson and Xiaobing Liu and Łukasz Kaiser and Stephan Gouws and Yoshikiyo Kato and Taku Kudo and Hideto Kazawa and Keith Stevens and George Kurian and Nishant Patil and Wei Wang and Cliff Young and Jason Smith and Jason Riesa and Alex Rudnick and Oriol Vinyals and Greg Corrado and Macduff Hughes and Jeffrey Dean},
      year={2016},
      eprint={1609.08144},
      archivePrefix={arXiv},
      primaryClass={cs.CL},
      url={https://arxiv.org/abs/1609.08144}, 
}

@article{remy-etal-2024-transtokenization,
  title={Trans-tokenization and cross-lingual vocabulary transfers: Language adaptation of LLMs for low-resource NLP},
  author={Remy, Fran{\c{c}}ois and Delobelle, Pieter and Avetisyan, Hayastan and Khabibullina, Alfiya and de Lhoneux, Miryam and Demeester, Thomas},
  journal={arXiv preprint arXiv:2408.04303},
  year={2024},
}

@inproceedings{lian-2025-scaffold,
  title={Scaffold-bpe: Enhancing byte pair encoding for large language models with simple and effective scaffold token removal},
  author={Lian, Haoran and Xiong, Yizhe and Niu, Jianwei and Mo, Shasha and Su, Zhenpeng and Lin, Zijia and Chen, Hui and Han, Jungong and Ding, Guiguang},
  booktitle={Proceedings of the AAAI Conference on Artificial Intelligence},
  volume={39},
  number={23},
  pages={24539--24548},
  year={2025},
}

@inproceedings{pilana-liyanage-yvon-2026-adaptbpe,
    title = "{A}dapt{BPE}: From General Purpose to Specialized Tokenizers",
    author = "Pilana Liyanage, Vijini  and
      Yvon, Fran{\c{c}}ois",
    editor = "Demberg, Vera  and
      Inui, Kentaro  and
      Marquez, Llu{\'i}s",
    booktitle = "Proceedings of the 19th Conference of the {E}uropean Chapter of the {A}ssociation for {C}omputational {L}inguistics (Volume 1: Long Papers)",
    month = mar,
    year = "2026",
    address = "Rabat, Morocco",
    publisher = "Association for Computational Linguistics",
    url = "https://aclanthology.org/2026.eacl-long.119/",
    doi = "10.18653/v1/2026.eacl-long.119",
    pages = "2607--2620",
    ISBN = "979-8-89176-380-7"
}

@inproceedings{sennrich-etal-2016-neural,
    title = "Neural Machine Translation of Rare Words with Subword Units",
    author = "Sennrich, Rico  and
      Haddow, Barry  and
      Birch, Alexandra",
    editor = "Erk, Katrin  and
      Smith, Noah A.",
    booktitle = "Proceedings of the 54th Annual Meeting of the Association for Computational Linguistics (Volume 1: Long Papers)",
    month = aug,
    year = "2016",
    address = "Berlin, Germany",
    publisher = "Association for Computational Linguistics",
    url = "https://aclanthology.org/P16-1162/",
    doi = "10.18653/v1/P16-1162",
    pages = "1715--1725"
}

@inproceedings{kudo-2018-subword,
    title = "Subword Regularization: Improving Neural Network Translation Models with Multiple Subword Candidates",
    author = "Kudo, Taku",
    editor = "Gurevych, Iryna  and
      Miyao, Yusuke",
    booktitle = "Proceedings of the 56th Annual Meeting of the Association for Computational Linguistics (Volume 1: Long Papers)",
    month = jul,
    year = "2018",
    address = "Melbourne, Australia",
    publisher = "Association for Computational Linguistics",
    url = "https://aclanthology.org/P18-1007/",
    doi = "10.18653/v1/P18-1007",
    pages = "66--75"
}

@inproceedings{sachidananda-etal-2021-efficient,
    title = "Efficient Domain Adaptation of Language Models via Adaptive Tokenization",
    author = "Sachidananda, Vin  and
      Kessler, Jason  and
      Lai, Yi-An",
    editor = "Moosavi, Nafise Sadat  and
      Gurevych, Iryna  and
      Fan, Angela  and
      Wolf, Thomas  and
      Hou, Yufang  and
      Marasovi{\'c}, Ana  and
      Ravi, Sujith",
    booktitle = "Proceedings of the Second Workshop on Simple and Efficient Natural Language Processing",
    month = nov,
    year = "2021",
    address = "Virtual",
    publisher = "Association for Computational Linguistics",
    url = "https://aclanthology.org/2021.sustainlp-1.16/",
    doi = "10.18653/v1/2021.sustainlp-1.16",
    pages = "155--165"
}

@inproceedings{minixhofer-etal-2022-wechsel,
    title = "{WECHSEL}: Effective initialization of subword embeddings for cross-lingual transfer of monolingual language models",
    author = "Minixhofer, Benjamin  and
      Paischer, Fabian  and
      Rekabsaz, Navid",
    editor = "Carpuat, Marine  and
      de Marneffe, Marie-Catherine  and
      Meza Ruiz, Ivan Vladimir",
    booktitle = "Proceedings of the 2022 Conference of the North American Chapter of the Association for Computational Linguistics: Human Language Technologies",
    month = jul,
    year = "2022",
    address = "Seattle, United States",
    publisher = "Association for Computational Linguistics",
    url = "https://aclanthology.org/2022.naacl-main.293/",
    doi = "10.18653/v1/2022.naacl-main.293",
    pages = "3992--4006"
}

@inproceedings{zouhar-etal-2023-tokenization,
    title = "Tokenization and the Noiseless Channel",
    author = "Zouhar, Vil{\'e}m  and
      Meister, Clara  and
      Gastaldi, Juan  and
      Du, Li  and
      Sachan, Mrinmaya  and
      Cotterell, Ryan",
    editor = "Rogers, Anna  and
      Boyd-Graber, Jordan  and
      Okazaki, Naoaki",
    booktitle = "Proceedings of the 61st Annual Meeting of the Association for Computational Linguistics (Volume 1: Long Papers)",
    month = jul,
    year = "2023",
    address = "Toronto, Canada",
    publisher = "Association for Computational Linguistics",
    url = "https://aclanthology.org/2023.acl-long.284/",
    doi = "10.18653/v1/2023.acl-long.284",
    pages = "5184--5207"
}

@inproceedings{xu-etal-2021-vocabulary,
    title = "Vocabulary Learning via Optimal Transport for Neural Machine Translation",
    author = "Xu, Jingjing  and
      Zhou, Hao  and
      Gan, Chun  and
      Zheng, Zaixiang  and
      Li, Lei",
    editor = "Zong, Chengqing  and
      Xia, Fei  and
      Li, Wenjie  and
      Navigli, Roberto",
    booktitle = "Proceedings of the 59th Annual Meeting of the Association for Computational Linguistics and the 11th International Joint Conference on Natural Language Processing (Volume 1: Long Papers)",
    month = aug,
    year = "2021",
    address = "Online",
    publisher = "Association for Computational Linguistics",
    url = "https://aclanthology.org/2021.acl-long.571/",
    doi = "10.18653/v1/2021.acl-long.571",
    pages = "7361--7373"
}

@inproceedings{bostrom-durrett-2020-byte,
    title = "Byte Pair Encoding is Suboptimal for Language Model Pretraining",
    author = "Bostrom, Kaj  and
      Durrett, Greg",
    editor = "Cohn, Trevor  and
      He, Yulan  and
      Liu, Yang",
    booktitle = "Findings of the Association for Computational Linguistics: EMNLP 2020",
    month = nov,
    year = "2020",
    address = "Online",
    publisher = "Association for Computational Linguistics",
    url = "https://aclanthology.org/2020.findings-emnlp.414/",
    doi = "10.18653/v1/2020.findings-emnlp.414",
    pages = "4617--4624"
}

@inproceedings{bandarkar-etal-2024-belebele,
    title = "The Belebele Benchmark: a Parallel Reading Comprehension Dataset in 122 Language Variants",
    author = "Bandarkar, Lucas  and
      Liang, Davis  and
      Muller, Benjamin  and
      Artetxe, Mikel  and
      Shukla, Satya Narayan  and
      Husa, Donald  and
      Goyal, Naman  and
      Krishnan, Abhinandan  and
      Zettlemoyer, Luke  and
      Khabsa, Madian",
    editor = "Ku, Lun-Wei  and
      Martins, Andre  and
      Srikumar, Vivek",
    booktitle = "Proceedings of the 62nd Annual Meeting of the Association for Computational Linguistics (Volume 1: Long Papers)",
    month = aug,
    year = "2024",
    address = "Bangkok, Thailand",
    publisher = "Association for Computational Linguistics",
    url = "https://aclanthology.org/2024.acl-long.44/",
    doi = "10.18653/v1/2024.acl-long.44",
    pages = "749--775"
}

@inproceedings{anikina-etal-2025-rigorous,
    title = "A Rigorous Evaluation of {LLM} Data Generation Strategies for Low-Resource Languages",
    author = "Anikina, Tatiana  and
      Cegin, Jan  and
      Simko, Jakub  and
      Ostermann, Simon",
    editor = "Christodoulopoulos, Christos  and
      Chakraborty, Tanmoy  and
      Rose, Carolyn  and
      Peng, Violet",
    booktitle = "Proceedings of the 2025 Conference on Empirical Methods in Natural Language Processing",
    month = nov,
    year = "2025",
    address = "Suzhou, China",
    publisher = "Association for Computational Linguistics",
    url = "https://aclanthology.org/2025.emnlp-main.418/",
    doi = "10.18653/v1/2025.emnlp-main.418",
    pages = "8282--8303",
    ISBN = "979-8-89176-332-6"
}

@inproceedings{luo-etal-2025-survey,
    title = "A Survey on Efficient Large Language Model Training: From Data-centric Perspectives",
    author = "Luo, Junyu  and
      Wu, Bohan  and
      Luo, Xiao  and
      Xiao, Zhiping  and
      Jin, Yiqiao  and
      Tu, Rong-Cheng  and
      Yin, Nan  and
      Wang, Yifan  and
      Yuan, Jingyang  and
      Ju, Wei  and
      Zhang, Ming",
    editor = "Che, Wanxiang  and
      Nabende, Joyce  and
      Shutova, Ekaterina  and
      Pilehvar, Mohammad Taher",
    booktitle = "Proceedings of the 63rd Annual Meeting of the Association for Computational Linguistics (Volume 1: Long Papers)",
    month = jul,
    year = "2025",
    address = "Vienna, Austria",
    publisher = "Association for Computational Linguistics",
    url = "https://aclanthology.org/2025.acl-long.1493/",
    doi = "10.18653/v1/2025.acl-long.1493",
    pages = "30904--30920",
    ISBN = "979-8-89176-251-0"
}

@inproceedings{kudo-richardson-2018-sentencepiece,
    title = "{S}entence{P}iece: A simple and language independent subword tokenizer and detokenizer for Neural Text Processing",
    author = "Kudo, Taku  and
      Richardson, John",
    editor = "Blanco, Eduardo  and
      Lu, Wei",
    booktitle = "Proceedings of the 2018 Conference on Empirical Methods in Natural Language Processing: System Demonstrations",
    month = nov,
    year = "2018",
    address = "Brussels, Belgium",
    publisher = "Association for Computational Linguistics",
    url = "https://aclanthology.org/D18-2012/",
    doi = "10.18653/v1/D18-2012",
    pages = "66--71"
}

@inproceedings{tiedemann-2012-parallel,
    title = "Parallel Data, Tools and Interfaces in {OPUS}",
    author = {Tiedemann, J{\"o}rg},
    editor = "Calzolari, Nicoletta  and
      Choukri, Khalid  and
      Declerck, Thierry  and
      Do{\u{g}}an, Mehmet U{\u{g}}ur  and
      Maegaard, Bente  and
      Mariani, Joseph  and
      Moreno, Asuncion  and
      Odijk, Jan  and
      Piperidis, Stelios",
    booktitle = "Proceedings of the Eighth International Conference on Language Resources and Evaluation ({LREC}'12)",
    month = may,
    year = "2012",
    address = "Istanbul, Turkey",
    publisher = "European Language Resources Association (ELRA)",
    url = "https://aclanthology.org/L12-1246/",
    pages = "2214--2218"
}

@inproceedings{leitner-etal-2020-dataset,
    title = "A Dataset of {G}erman Legal Documents for Named Entity Recognition",
    author = "Leitner, Elena  and
      Rehm, Georg  and
      Moreno-Schneider, Julian",
    editor = "Calzolari, Nicoletta  and
      B{\'e}chet, Fr{\'e}d{\'e}ric  and
      Blache, Philippe  and
      Choukri, Khalid  and
      Cieri, Christopher  and
      Declerck, Thierry  and
      Goggi, Sara  and
      Isahara, Hitoshi  and
      Maegaard, Bente  and
      Mariani, Joseph  and
      Mazo, H{\'e}l{\`e}ne  and
      Moreno, Asuncion  and
      Odijk, Jan  and
      Piperidis, Stelios",
    booktitle = "Proceedings of the Twelfth Language Resources and Evaluation Conference",
    month = may,
    year = "2020",
    address = "Marseille, France",
    publisher = "European Language Resources Association",
    url = "https://aclanthology.org/2020.lrec-1.551/",
    pages = "4478--4485",
    language = "eng",
    ISBN = "979-10-95546-34-4"
}

@inproceedings{papineni-etal-2002-bleu,
    title = "{B}leu: a Method for Automatic Evaluation of Machine Translation",
    author = "Papineni, Kishore  and
      Roukos, Salim  and
      Ward, Todd  and
      Zhu, Wei-Jing",
    editor = "Isabelle, Pierre  and
      Charniak, Eugene  and
      Lin, Dekang",
    booktitle = "Proceedings of the 40th Annual Meeting of the Association for Computational Linguistics",
    month = jul,
    year = "2002",
    address = "Philadelphia, Pennsylvania, USA",
    publisher = "Association for Computational Linguistics",
    url = "https://aclanthology.org/P02-1040/",
    doi = "10.3115/1073083.1073135",
    pages = "311--318"
}

@inproceedings{post-2018-call,
    title = "A Call for Clarity in Reporting {BLEU} Scores",
    author = "Post, Matt",
    editor = "Bojar, Ond{\v{r}}ej  and
      Chatterjee, Rajen  and
      Federmann, Christian  and
      Fishel, Mark  and
      Graham, Yvette  and
      Haddow, Barry  and
      Huck, Matthias  and
      Yepes, Antonio Jimeno  and
      Koehn, Philipp  and
      Monz, Christof  and
      Negri, Matteo  and
      N{\'e}v{\'e}ol, Aur{\'e}lie  and
      Neves, Mariana  and
      Post, Matt  and
      Specia, Lucia  and
      Turchi, Marco  and
      Verspoor, Karin",
    booktitle = "Proceedings of the Third Conference on Machine Translation: Research Papers",
    month = oct,
    year = "2018",
    address = "Brussels, Belgium",
    publisher = "Association for Computational Linguistics",
    url = "https://aclanthology.org/W18-6319/",
    doi = "10.18653/v1/W18-6319",
    pages = "186--191"
}

\appendix

\section{Appendix}
\label{sec:appendix}

\subsection{Merge selection Algorithm}\label{sec:comp-scaffold-nml}
\done\todo{Use the same notation as in the main text}

Algorithm~\ref{alg:comp-scaffold-nml} describe the main procedure for MDL-COMP; for the other versions it suffices to replace the computation of the complexity term in line~\ref{line:complexity} with the appropriate score, and also the computation of CRIT (for VOLT). For the computation in line~\ref{line:loglik}, we use the fact that for a given corpus state with unigram counts $\{n_t\}$ and total token count $T=\sum_t n_t$, the log-likelihood is $\ell(\{n_t\}) = \sum_t n_t \log n_t - T \log T$. The computations in lines \ref{line:update_logl}-\ref{line:update_comp} are incremental, as the difference in tokenization of $\corpus^t$ and $\corpus^u$ are minimal, and yield unigram counts that only differ for 3 units, and parameter counts that only differ by 1. For completeness, we recall the scaffolding procedure in Algorithm~\ref{alg:scaffolding}.

\begin{algorithm}[htbp]
\caption{Complexity-based Merge Refinement}
\label{alg:comp-scaffold-nml}
\begin{algorithmic}[1]
\Require Merge list $\boldsymbol{\mu} = [\mu_0, \dots, \mu_M]$
\Require corpus $\corpus$, initially tokenized in $U$ units\label{line:init}
\State \textbf{Phase I: Greedy Merge List Filtering}
\State $\boldsymbol{\mu}_A \gets \emptyset$ \Comment{Accepted merges}
\State $\corpus^t \gets \corpus$
\State $\text{LL}(\corpus^t) \gets \sum_t n_t \log n_t - T \log T$ \label{line:loglik}
\State $\text{COMP}(\corpus^t) \gets 0.5 \cdot (U - 1) \cdot \log(S)$ \label{line:complexity}
\State $\text{CRIT}(\corpus^t) \gets \text{COMP}(\corpus^t) -\text{LL}(\corpus^t)$

\For{$i = 0$ \textbf{to} $M$} 
    \State $\corpus^u \gets \operatorname{apply}(\mu_i, \corpus^t)$
    \State Compute $\text{LL}(\corpus^u)$ \label{line:update_logl}
    \State Compute $\text{COMP}(\corpus^u)$ \label{line:update_comp}
    \State Compute $\text{CRIT}(\corpus^u)$ \label{line:update_crit}
    \State $\Delta \text{CRIT} \gets \text{CRIT}(\corpus^u) - \text{CRIT}(\corpus^t)$
    \If{$\Delta \text{CRIT} < 0$} 
        \State $\boldsymbol{\mu}_A \gets \boldsymbol{\mu}_A \cup \{\mu_i\}$
        \State $\corpus^t \gets \corpus^u$
    \EndIf
\EndFor
\State \Return $\boldsymbol{\mu}_\text{A}$
\end{algorithmic}
\end{algorithm}

\begin{algorithm}[htbp]
\caption{Scaffolding procedure}
\label{alg:scaffolding}
\begin{algorithmic}[1]
  \Require Merge list $\boldsymbol{\mu} = [\mu_0, \dots, \mu_M]$
  \Require Accepted merge list $\boldsymbol{\mu}_{\text{A}}$
  \Require corpus $\corpus$, initially tokenized in $U$ units
\State \textbf{Phase II: Scaffolding Refinement}
\State $\corpus^t \gets \operatorname{apply}(\boldsymbol{\mu}_{\text{A}},\corpus)$
\State Compute unigram frequencies: \\ $\quad \operatorname{Freq}(\mu, \corpus^t), \forall \mu \in \boldsymbol{\mu}_{\text{A}}$
\State Compute bigram frequencies: \\ $\quad \operatorname{Freq}(\mu\mu', \corpus^t), \forall (\mu, \mu') \in \boldsymbol{\mu}_{\text{A}} \times \boldsymbol{\mu}_{\text{A}} $
\State $\mu_p \gets \arg \min_{\mu \in \boldsymbol{\mu}_\text{A}} \operatorname{Freq}(\mu, \corpus^t)$ 
\State $\mu_q \gets \arg \max_{\mu = (x,y) \in \boldsymbol{\mu}_\text{R}} \operatorname{Freq}(xy, \corpus^t)$ 
\While{$\operatorname{Freq}(\mu_p, \corpus^t) < \operatorname{Freq}(\mu_q, \corpus^t)$}
    \State Make $\mu_p$ a virtual merge
    \State $\corpus^t \gets \operatorname{unapply}(\mu_p, \corpus^t)$
    \State $\corpus^t \gets \operatorname{apply}(\mu_q, \corpus^t)$
    \State $\boldsymbol{\mu}_{\text{A}} \gets \boldsymbol{\mu}_{\text{A}} + [\mu_q]$
    \State $\boldsymbol{\mu}_{\text{R}} \gets \boldsymbol{\mu}_{\text{R}} - [\mu_q]$
    \State Update unigram and bigram frequencies
    \State $\mu_p \gets \arg \min_{\mu \in \boldsymbol{\mu}_\text{A}} \operatorname{Freq}(\mu, \corpus^t)$ 
\State $\mu_q \gets \arg \max_{\mu = (x,y) \in \boldsymbol{\mu}_\text{R}} \operatorname{Freq}(xy, \corpus^t)$ 
\EndWhile

\State \Return $\boldsymbol{\mu}_\text{A}$
\end{algorithmic}
\end{algorithm}

\subsection{Merge Retention Patterns Across Filtering Strategies}

\begin{figure*}[htbp]
\centering
\includegraphics[width=1.0\textwidth]{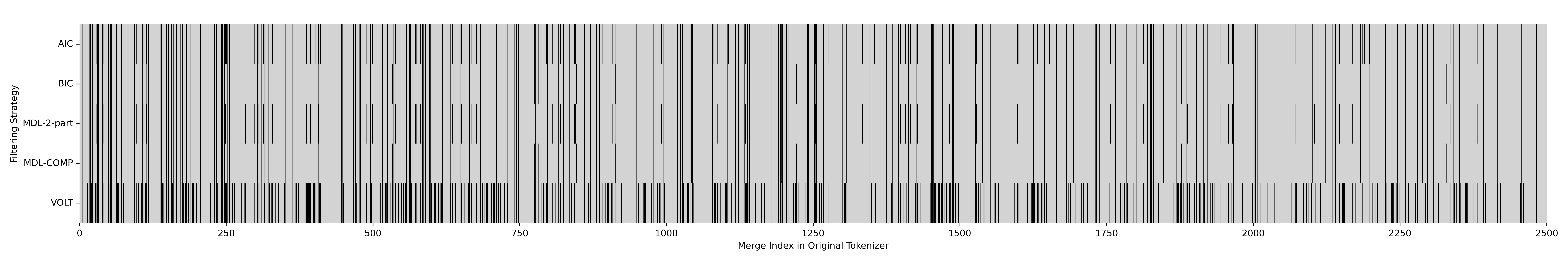}
\caption{\small {Retention map of the first 2.5k merges from the original tokenizer using 10k original EMEA English data.}}
\label{fig:merge_retention}
\end{figure*}

Figure~\ref{fig:merge_retention} shows how different filtering strategies retain or skip merges. Each row corresponds to a filtering strategy and each column to a merge index in the original tokenizer. Dark cells indicate retained merges, while light cells indicate removed merges.

The figure shows that most strategies consistently retain early merges, while differences become more pronounced for later merges. We also observe that VOLT retains a larger proportion of merges overall, as reflected by denser dark regions compared to other strategies.

\subsection{French-English Machine Translation Results}
\label{app:fre-eng-results}

Table~\ref{tab:MT-fre-eng} reports BLEU scores for French-English translation on Flores-200. The overall trends are consistent with the Sinhala-English setting discussed in the main text: VOLT and MDL-COMP outperform AIC and BIC among the selection-based approaches, scaffolding consistently improves performance, and synthetic data generally yields additional gains.

\begin{table*}[h!]
\centering
\resizebox{\textwidth}{!}{%
\begin{tabular}{lcccccccccccc}
\toprule
\multirow{2}{*}{\textbf{Strategy}}
& \multicolumn{4}{c}{\textbf{orig}} 
& \multicolumn{4}{c}{\textbf{gen}} 
& \multicolumn{4}{c}{\textbf{orig+gen}} \\
\cmidrule(lr){2-5} \cmidrule(lr){6-9} \cmidrule(lr){10-13}
 & 5k+5k & 10k+10k & 20k+20k & 50k+50k 
 & 5k+5k & 10k+10k & 20k+20k & 50k+50k  
 & 5k+5k & 10k+10k & 20k+20k & 50k+50k  \\
\midrule

Adapt-BPE
& 45.32 (30k) & 45.32 (30k)  & 45.61 (30k) & 45.73 (30k) 
& 42.12 (30k) & 42.57 (30k)  & 42.91 (30k) & 43.77 (30k) 
& 45.35 (30k) & 45.43 (30k)  & 45.77 (30k) & 45.77 (30k)\\

AIC
& 34.02 (6.6k) & 34.06 (7.5k) & 34.06 (8.0k) & 34.06 (8.5k)
& 31.92 (7.7k) & 32.23 (8.3k) & 32.08 (8.8k) & 32.98 (10.0k)
& 34.10 (8.1k) & 34.10 (8.5k) & 34.56 (9.8k) & 34.59 (10.4k) \\

\quad  + Scaffolding
& 34.53 (6.6k) & 34.58 (7.5k) & 34.58 (8.0k) & 34.60 (8.5k)
& 32.02 (7.7k) & 32.29 (8.3k) & 32.11 (8.8k) & 32.98 (10.0k)
& 34.56 (8.1k) & 34.56 (8.5k) & 34.67 (9.8k) & 34.80 (10.4k) \\

BIC 
& 29.02 (3.4k) & 29.08 (4.2k) & 30.20 (4.9k) & 30.88 (5.5k)
& 27.56 (4.0k) & 27.57 (4.8k) & 27.67 (5.3k) & 27.88 (5.7k)
& 29.60 (5.2k) & 29.66 (5.8k) & 30.09 (6.9k) & 30.15 (7.4k)\\

\quad + Scaffolding
& 29.12 (3.4k) & 29.17 4.2k) & 30.21 (4.9k) & 31.05 (5.5k)
& 28.06 (4.0k) & 28.19 (4.8k) & 28.29 (5.3k) & 28.53 (5.7k)
& 29.69 (5.2k) & 29.69 (5.8k) & 30.13 (6.9k) & 30.25 (7.4k)\\

MDL-2-part 
& 30.92 (3.9k) & 30.92 (4.2k) & 31.21 (4.4k) & 31.25 (4.6k)
& 29.80 (5.1k) & 29.83 (5.1k) & 29.85 (5.4k) & 29.85 (6.1k)
& 31.21 (6.0k) & 31.14 (6.6k) & 31.34 (7.9k) & 31.95 (7.8k)\\

\quad  + Scaffolding
& 31.02 (3.9k) & 31.11(4.2k) & 31.29 (4.4k) & 31.34 (4.6k)
& 30.09 (5.1k) & 30.13 (5.1k) & 30.15 (5.4k) & 30.35 (6.1k)
& 32.12 (6.0k) & 32.14 (6.6k) & 32.59 (7.9k) & 33.01 (7.8k)\\

MDL-COMP 
& 31.54 (4.4k) & 31.45 (4.8k) & 31.67 (5.7k) & 31.68 (5.2k)
& 32.01 (5.7k) & 32.24 (6.1k) & 32.45 (6.9k) & 32.46 (7.2k)
& 33.01 (6.2k) & 33.09 (6.8k) & 33.09 (7.9k) & 33.09 (8.3k)\\

\quad  + Scaffolding
& 31.64 (4.4k) & 31.72 (4.8k) & 31.75 (5.7k) & 32.03 (5.2k)
& 32.09 (5.7k) & 32.33 (6.1k) & 32.78 (6.9k) & 32.98 (7.2k)
& 33.55 (6.2k) & 33.67 (6.8) & 33.69 (7.9k) & 33.69 (8.3k)\\

VOLT
& 33.54 (7.2k) & 33.45 (8.2k) & 33.67 (8.1k) & 33.68 (8.1k)
& 32.01 (8.7k) & 32.24 (9.2k) & 32.45 (9.0k) & 32.46 (9.8k)
& 34.01 (10.5k) & 34.09 (10.1k) & 34.09 (11.3k) & 34.09 (11.3k)\\

\quad + Scaffolding
& 35.14 (7.2k) & 35.39 (8.2k) & 35.78 (8.1k) & 35.99 (8.1k)
& 35.12 (8.7k) & 35.14 (9.2k) & 35.44 (9.0k) & 35.56 (9.8k)
& 36.07 (10.5k) & 36.07 (10.1k) & 36.19 (11.3k) & 36.65 (11.3k)\\

\bottomrule
\end{tabular}
}
\captionsetup{font={tiny}, labelfont={tiny}}
\caption{MT results for fre-eng Flores-200 data (BLEU). Using the full tokenizer yields BLEU=45.6.}
\label{tab:MT-fre-eng}
\end{table*}

\subsection{Machine Translation Details}

\paragraph{Model}
We use \texttt{LLaMA-3-8B} for all translation experiments.

\paragraph{Prompting Setup}
We use a zero-shot prompting format:

\begin{quote}
Translate the following sentence from French to English:\\
Source: \{sentence\}\\
Target:
\end{quote}

For Sinhala-English, we use a similar format:
\begin{quote}
Translate the following sentence from Sinhala to English:\\
Source: \{sentence\}\\
Target:
\end{quote}

\paragraph{Decoding}
We use greedy decoding with a max length of 128 tokens.\done\todo{Temperature is irrelevant for greedy decoding}


\subsection{Reading Comprehension Evaluation} \label{ann:reading_comprehension}

\paragraph{Model Input Format}
Each Belebele instance consists of:
\begin{itemize}
    \item a passage
    \item a question
    \item multiple-choice answers
\end{itemize}

We format the input as:
\begin{quote}
Passage: \{passage\}\\
Question: \{question\}\\
Options:\\
A. \{A\}\\
B. \{B\}\\
C. \{C\}\\
D. \{D\}\\
Answer:
\end{quote}

\paragraph{Prediction}
The model is prompted to output only the option letter (A/B/C/D). The final answer is extracted via exact match.

\subsection{Appendix: Corpus Statistics}
\label{app:corpus-stats}

\begin{table}[h]
\centering
\small
\begin{tabular}{lcc}
\toprule
\textbf{Corpus} & \textbf{Language(s)} & \textbf{\# Sentences}  \\
\midrule
EMEA & eng, fra & 1.1M  \\
LER  & deu & 53.4k  \\
Opus-swh & eng, swh & 94.6k  \\
Opus-sin & eng, sin & 3.2M  \\
\bottomrule
\end{tabular}
\caption{Statistics of domain-specific and multilingual corpora used in experiments.}
\label{tab:corpus-stats}
\end{table}

\todos{}

\end{document}